\documentclass[letterpaper]{article} 
\usepackage{aaai23}  
\usepackage{times}  
\usepackage{helvet}  
\usepackage{courier}  
\usepackage[hyphens]{url}  
\usepackage{graphicx} 
\usepackage{natbib}  
\usepackage{caption} 
\usepackage{amsmath,amssymb,amsfonts}
\usepackage{algorithm}
\usepackage{algorithmic}

\usepackage{xcolor}

\usepackage{tikz}
\usetikzlibrary{shapes,arrows,backgrounds,fit,shadows,positioning}
\usepackage{colortbl}
\newcommand{\sol}[1]{\mbox{\emph{#1}}}

\usepackage[normalem]{ulem} 

\newcommand{\Pre}{\mathit{Pre}}
\newcommand{\Eff}{\mathit{Eff}}
\newcommand{\true}{\mathit{true}}
\newcommand{\false}{\mathit{false}}

\newcommand{\remove}[1]{}

\newtheorem{example}{Example}

\usepackage{newfloat}
\usepackage{listings}
\DeclareCaptionStyle{ruled}{labelfont=normalfont,labelsep=colon,strut=off} 
\floatstyle{ruled}
\newfloat{listing}{tb}{lst}{}
\floatname{listing}{Listing}
\title{A Good Snowman is Hard to Plan}
\author {
    Miquel Bofill,\textsuperscript{\rm 1}
    Cristina Borralleras, \textsuperscript{\rm 2}
    Joan Espasa, \textsuperscript{\rm 3}
    Gerard Mart\'in,\textsuperscript{\rm 1}
    Gustavo Patow,\textsuperscript{\rm 1}
    Mateu~Villaret\textsuperscript{\rm 1}
}
\affiliations {
    \textsuperscript{\rm 1} Departament d'Inform\`atica, Matem\`atica Aplicada i Estad\'istica, Universitat de Girona, Spain\\
    \textsuperscript{\rm 2} Departament d'Enginyeries, Universitat de Vic - Universitat Central de Catalunya, Spain\\
    \textsuperscript{\rm 3} School of Computer Science, University of St Andrews, UK\\
    miquel.bofill@udg.edu, cristina.borralleras@uvic.cat, jea20@st-andrews.ac.uk, gerardmartinteixidor@gmail.com, gustavo.patow@udg.edu, mateu.villaret@udg.edu
}

\begin{document}

\maketitle

\begin{abstract}
In this work we face a challenging puzzle video game: \emph{A~Good Snowman is Hard to Build}.
The objective of the game is to build snowmen by  moving and stacking snowballs on a discrete grid.  
For the sake of player engagement with the game, it is interesting to avoid that a player finds a much easier solution than the one the designer expected. 
Therefore, having tools that are able to certify the optimality of solutions is crucial.

Although the game can be stated as a planning problem and can be naturally modelled in PDDL, we show that a direct translation to SAT clearly outperforms off-the-shelf state-of-the-art planners. As we show, this is mainly due to the fact that reachability properties can be easily modelled in SAT, allowing for shorter plans, whereas using axioms to express a reachability derived predicate in PDDL does not result in any significant reduction of solving time  with the considered planners. 
We deal with a set of 51 levels, both original and crafted, solving 43 and with 8 challenging instances still remaining to be solved.
\end{abstract}

\section{Introduction}

During the design of video games, one of the most important aspects to consider is the difficulty of each level. A common problem while designing the levels is the possibility of ignoring a possible solution to the level that is much easier than the ones the designer had initially in mind~\cite{Silli2010}. These unplanned solutions often are not desired because they break the difficulty slope, making the game uninteresting. 
To this end, it is of crucial interest to have a tool able to provide (certified) optimal solutions for given scenarios of the game.

In this work we focus on finding optimal solutions for a discrete time and space puzzle. In particular we consider the PSPACE-complete game \emph{A Good Snowman is Hard to Build}~\cite{snowmangame,snowmanspace}, where a character
has to move snowballs to build snowmen. 
The game has some resemblance with the widely studied Sokoban~\cite{sokopspace} game, because the character pushes balls (instead of boxes) in a matrix-like maze with some obstacles. 
However, it generally gets more involved because not only scenario characteristics (positions of balls and snow) may change when moving snowballs, but the final positions of snowmen are not fixed, hence the goal is conceptually disjunctive, making the search space much bigger in comparison. Therefore, both the modelling and the increasing computational complexity makes the considered game  challenging.
In these kind of games, the difficulty of a level is sometimes measured using the number of times the character needs to move an object. There are two main reasons for this: first, moving the character from one location to another uses to be trivial because it consists on identifying if there is a path from one location to another; second, pushing an object may close free paths. 


\emph{A Good Snowman is Hard to Build} is naturally characterised as a planning problem, where we are asked to find a sequence of actions such as rolling balls on snow, and stacking or unstacking them, so that balls of the appropriate sizes are joined into snowmen. In the planning literature, most models for Sokoban have two actions: \emph{move}, for changing the location of the character, and \emph{push} for changing the location of a box by making the character push it. 
%
Some previous works devise methods to focus the search efforts on certain parts of the problem. In~\citet{IvankovicH15}, the authors show how a reachability derived predicate can be specified with axioms and the \emph{move} actions can be completely avoided in the model. The idea is to ensure with the reachability predicate that the pushing location is \emph{reachable} from the current character location. This results in shorter plans and in a significant reduction of the search space and hence in the time required to solve the instances. Similarly, a proposal to automatise the inference of axioms and to reformulate the problem accordingly is given in~\citet{MiuraF17}. This is done with success for Sokoban and  reachability, as shown by using axiom supporting model-based planners with Integer programming and Answer Set Programming technologies. 

In this paper, we start by providing a model for \emph{A Good Snowman is Hard to Build} written in PDDL~\cite{pddl}, the de-facto standard language for AI planners. This model contains actions to both \emph{move} the character and to \emph{push} snowballs, and we evaluate the performance of some state-of-the-art planners on it.
We also provide a PDDL model using  axioms for the reachability predicate, hence, only considering \emph{push} actions, which preconditions require that the location  where the action is performed is reachable from the location where the character was at the previous time step. Unfortunately, despite the detailed instructions for the planner from~\citet{IvankovicH15} we have not been able to install it, and the planner from~\citet{MiuraF17} was not available. Therefore, we use a state-of-the-art planner supporting derived predicates but without any specialized heuristics for them.
To fully assess the overhead of the \emph{move} actions, we also consider a ``what if'' scenario, where the character can do unsound teleports. That is, teleports that do not guarantee that there exists a valid path from source to destination (a sort of cheating). Interestingly, experiments show that this does not have a significant impact in solving time.

Next, we present an ad-hoc planning as SAT approach. 
Again, we then get rid of \emph{move} actions by encoding reachability. This results in a significant reduction of the time horizon and, accordingly, of the solving times, allowing to solve many more instances to optimality. 
By using the proposed planning as SAT approach with reachability, we are able to certify the minimum number of ball movements needed to solve most of the original levels of \emph{A Good Snowman is Hard to Build} and of the additional crafted instances that we also provide.

The main contribution of this paper is a new case study with a set of challenging PDDL instances (with and without derived predicates) and a detailed comparison of planning tools against planning as SAT methodology. 
For the planning as SAT approach we use the s-t-reachability encoding from~\citet{GebserJR14,DBLP:journals/corr/abs-2105-12908} and the folklore one based on neighbour counting. We  adapt these encodings to deal with a priori unknown source and target locations in a dynamic environment.

For the sake of reproducibility, the presented PDDL models and instances and the SAT encoding generators can be found in the supplementary material at \url{https://github.com/udg-lai/KEPS2023}. In addition, the full detailed experimental results are also included.
\section{The Game}

\emph{A Good Snowman Is Hard To Build} is a single-agent puzzle video game where the goal is to push snowballs in a maze to build some snowmen by stacking three balls of decreasing size. 

The game elements are the agent (i.e., the black character controlled by the player), the playable cells, which may or not contain snow, and the snowballs, which are initially distributed on the playable cells.
Snowballs have three possible sizes: small, medium and large.
As in Sokoban, the only allowed action is moving the agent in one of four directions. The results of \emph{moving} depend on the cells in front:

\begin{itemize}
\item \emph{Move}: When the agent walks into a free cell, he simply moves to that cell.
\item \emph{Roll}: When the agent moves into a cell with a single ball, and there is a free cell in front of the ball, the ball gets pushed and the agent occupies the cell previously occupied by the ball. If a ball is pushed into a snow cell, the snow disappears and the ball increases in size, up to a maximum. Still, the snow is always removed.
\item \emph{Push}: A ball can be pushed on a stack of balls if the size of the ball in the top is bigger than the one being pushed. Then, the agent occupies the cell previously occupied by the pushed ball like when rolling.
\item \emph{Pop}: Trying to move into a stack of balls will pop the topmost ball but will not change the location of the agent. This action can only happen if the ball falls directly into a cell without any ball.
\end{itemize}
Notice that, as in Sokoban, the agent is not allowed to pull balls.

\begin{figure}
\centering
\includegraphics[width=0.2\textwidth]{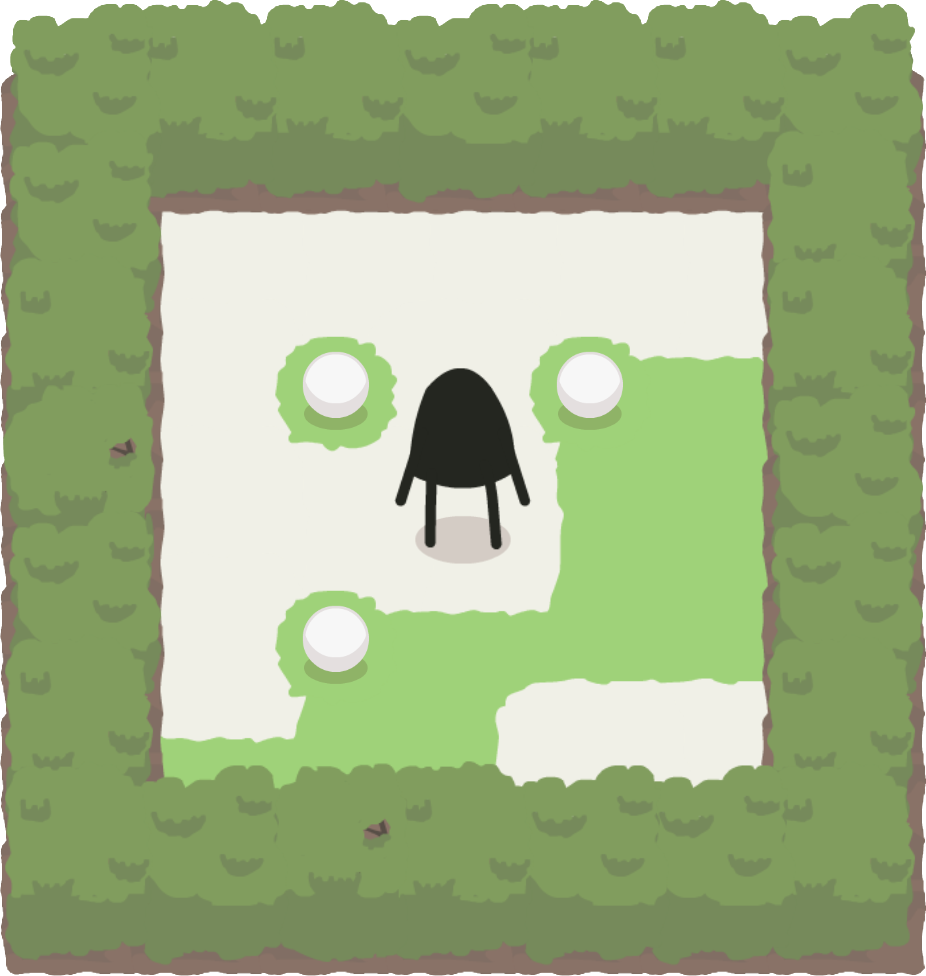}
\includegraphics[width=0.2\textwidth]{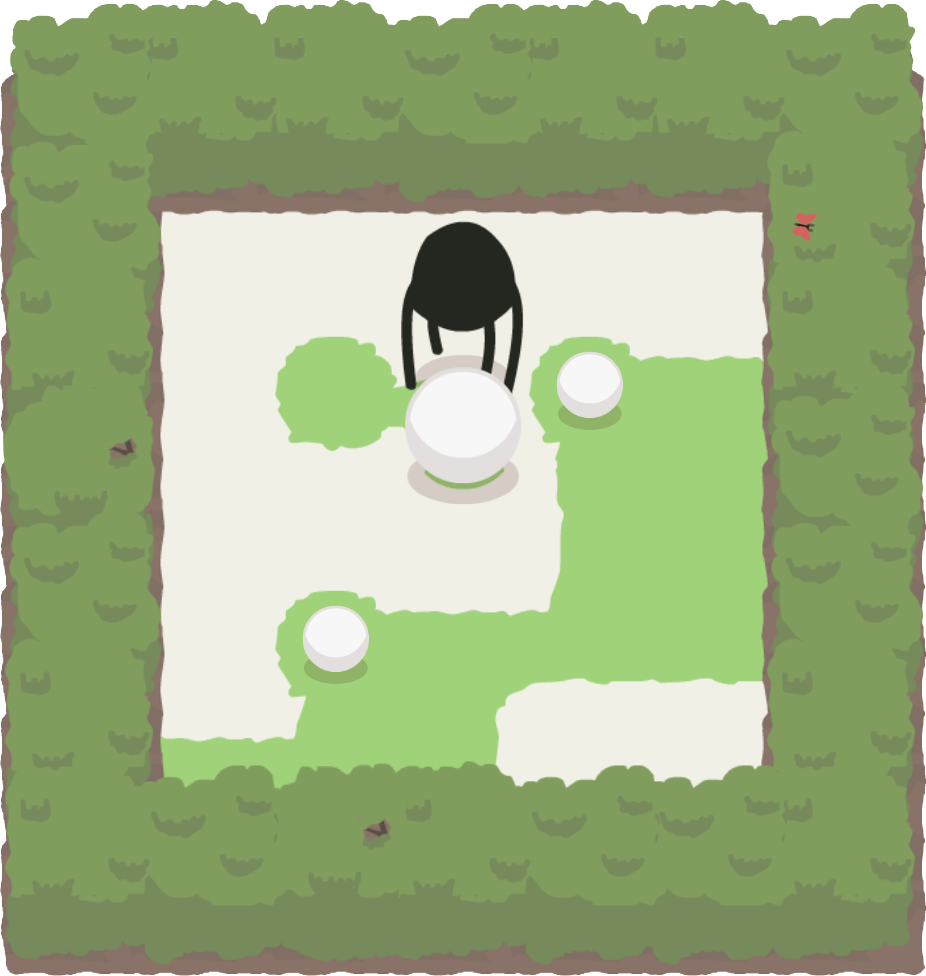}
\includegraphics[width=0.2\textwidth]{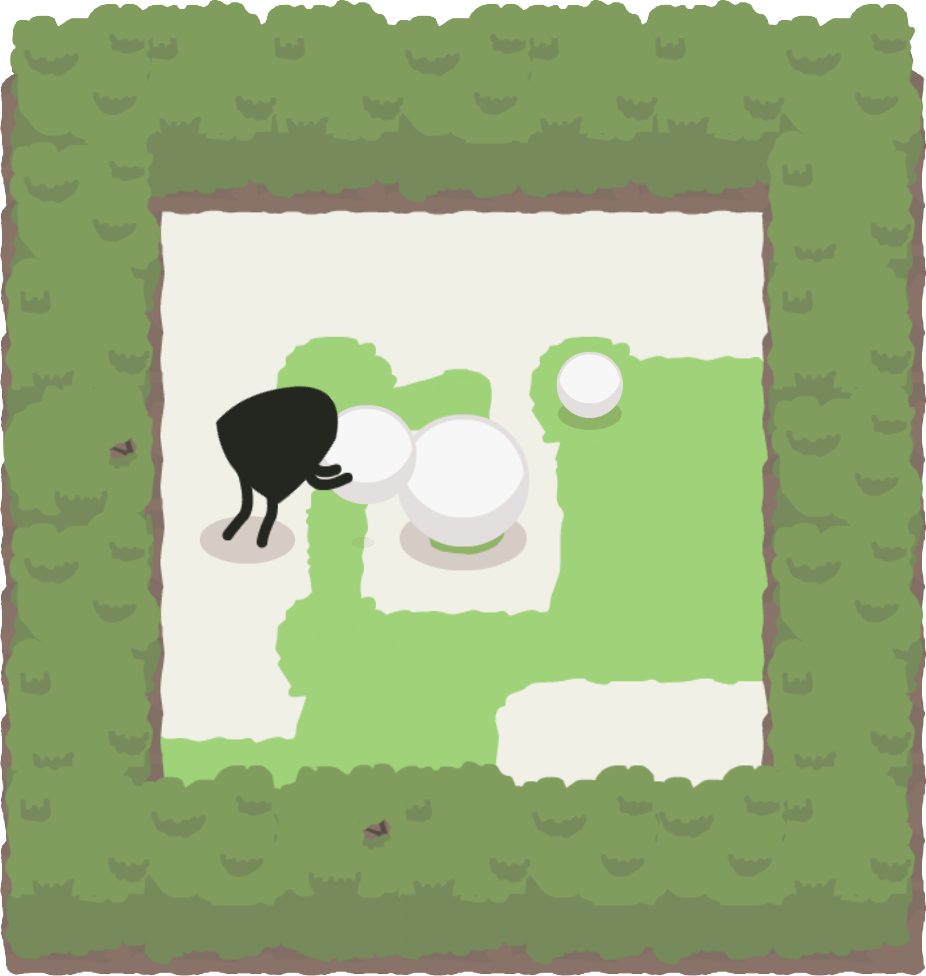}
\includegraphics[width=0.2\textwidth]{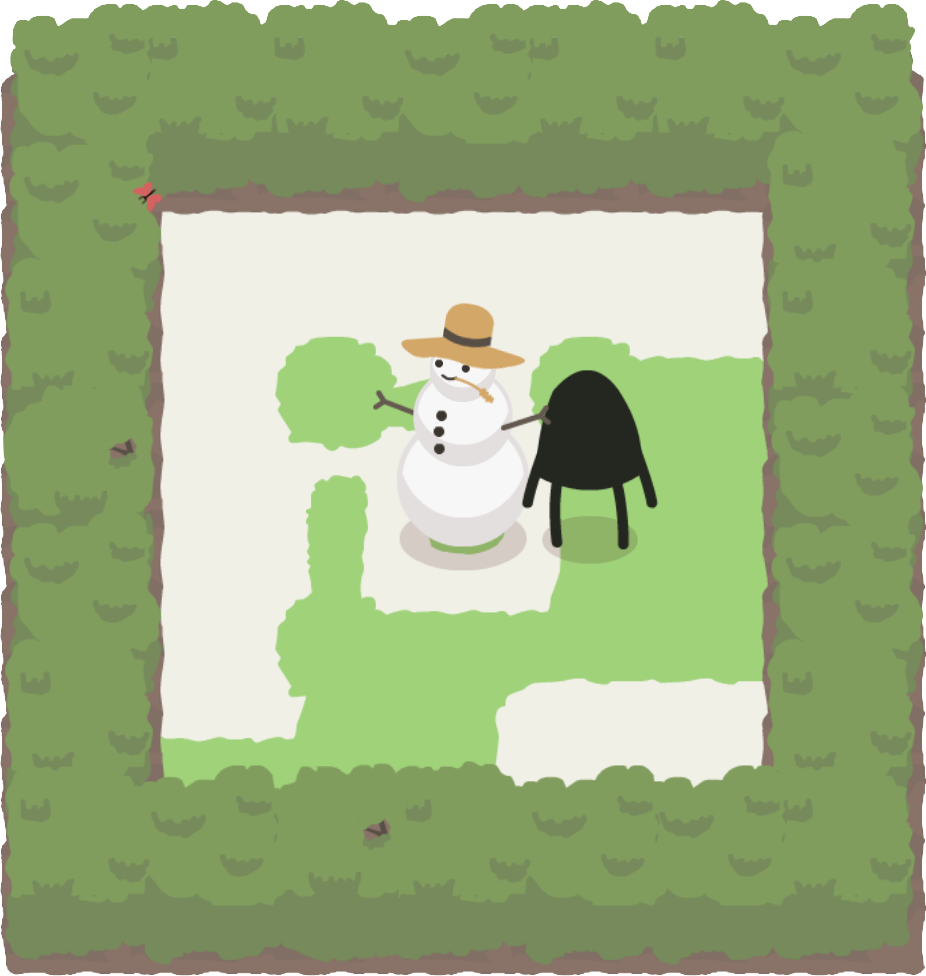}
\caption{\emph{Andy} level, showing the execution of the optimal solution \sol{lluRurDlldddrUluRuurrrdLulD}. Letters represent the direction of movement. Uppercase letters indicate ball movements.}\label{fig:solution_example}
\end{figure}

The goal of the game is to build snowmen composed by a pile of three balls of decreasing size. The scenarios of the game considered consist of three, six or nine snowballs, hence, one, two or three snowmen. Snowmen can be built anywhere. Figure~\ref{fig:solution_example} depicts an example of how to solve one of the levels of the game.

\section{PDDL Formulation}
Since the problem is very similar to Sokoban, we consider a PDDL formulation based on the Sokoban domain used in the 2008 and 2011 International Planning Competitions. 
However, when encoding it in PDDL there are various key differences, as the usage of quantifiers and conditional effects is essential. This is due to elements such as: disappearing snow, ball stacking and the disjunctive goal.
%
In this section we partially describe this PDDL formulation.\footnote{Full models and instances can be found at \url{https://github.com/udg-lai/KEPS2023}.}

\subsubsection*{Types, objects and fluents} 
The objects considered have the following types: \texttt{loc}, \texttt{dir} and \texttt{ball}. The \texttt{loc} type represents a grid cell location, which may contain the character or a stack of balls. In addition to these, a cell may also contain snow. The \texttt{dir} type enumerates the four directions the character and balls can move (\texttt{up}, \texttt{down}, \texttt{left}, \texttt{right}) and  the \texttt{ball} type defines all possible balls of the scenario, e.g., \texttt{ball1}, \texttt{ball2} and \texttt{ball3}. Note that when building two snowmen there will be six balls, whereas when building three snowmen there will be nine balls.
In order to model the current state, the following predicates are defined:
\begin{itemize}
    \item \texttt{(snow ?l\,-\,loc)}: location \texttt{l} is covered in snow.
    \item \texttt{(next ?from ?to\,-\,loc ?d\,-\,dir)}: locations \texttt{from} and \texttt{to} are are adjacent; location \texttt{to} is in direction \texttt{d} relative to \texttt{from}.
    \item \texttt{(occ ?l\,-\,loc)}: location \texttt{l} contains at least one ball.
    \item \texttt{(char\_at ?l\,-\,loc)}: character is at location \texttt{l}.
    \item \texttt{(ball\_at ?b\,-\,ball ?l\,-\,loc)}: ball \texttt{b} is at location \texttt{l}.
    \item \texttt{(ball\_size\_s ?b\,-\,ball)}: ball \texttt{b} has small size.
    \item \texttt{(ball\_size\_m ?b\,-\,ball)}: ball \texttt{b} has medium size.
    \item \texttt{(ball\_size\_l ?b\,-\,ball)}:  ball \texttt{b} has large size.
    \item \texttt{(goal)}: the goal is satisfied.
\end{itemize}


\subsubsection*{Actions}

\begin{figure}
\begin{center}
\includegraphics[width=0.3\linewidth]{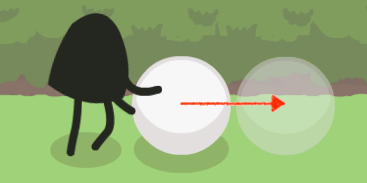}

\texttt{ppos}~~~~\texttt{from}~~~~~\texttt{to}~~~
\end{center}
\caption{Location parameters of action \texttt{move\_ball}.}
\label{fig:pddl_ppos_from_to}
\end{figure}

Three actions are defined: \texttt{move\_character}, \texttt{move\_ball} and \texttt{goal}.
As its name indicates,  action \texttt{move\_character} moves the character from its current location to some of its next adjacent locations, according to a direction. 



Action \texttt{move\_ball} moves a single ball. This action has as parameters a ball to be moved (\texttt{b}) and three aligned and adjacent locations (\texttt{ppos}, \texttt{from} and \texttt{to}). Figure~\ref{fig:pddl_ppos_from_to} illustrates the role of the three location parameters. 
With this action, ball \texttt{b} always is going to move, but the character may not. 

\lstdefinelanguage{PDDL}{
  keywords={and, or, exists, forall, assign, not, when, increase},
  keywordstyle=\color{blue}\bfseries,
  ndkeywords={action, parameters, precondition, effect},
  ndkeywordstyle=\color{mygreen}\bfseries,
  identifierstyle=\color{black},
  sensitive=false,
  comment=[l]{;},
  commentstyle=\color{purple}\ttfamily,
  stringstyle=\color{red}\ttfamily,
  morestring=[b]',
  morestring=[b]"
}
\definecolor{mygreen}{rgb}{0,0.6,0}
\lstset{
  basicstyle=\scriptsize\ttfamily,
  columns=fullflexible, keepspaces=true,
  tabsize=2, stepnumber=1,
  xleftmargin=2em
}
\lstset{
  basicstyle=\scriptsize\ttfamily,
  columns=fullflexible, keepspaces=true,
  numbers=left, tabsize=2, stepnumber=1}
  \begin{figure}
\begin{lstlisting}[escapechar=|, language=PDDL]
(:action move_ball
 :parameters 
    (?b - ball ?ppos ?from ?to - loc ?d - dir)
 :precondition (and
    (next ?ppos ?from ?d) (next ?from ?to ?d)|\label{line:1}|
    (ball_at ?b ?from)                       |\label{line:2}|
    (char_at ?ppos)                          |\label{line:3}|
    (forall (?o - ball)                      |\label{line:4}|
        (or
            (= ?o ?b) ; Ball != from ball action
            (or ; Implication
                (not (ball_at ?o ?from))
                (or ; Ball ?b smaller than ?o
                    (and
                        (ball_size_small ?b)
                        (ball_size_medium ?o))
                    (and
                        (ball_size_small ?b)
                        (ball_size_large ?o))
                    (and
                        (ball_size_medium ?b)
                        (ball_size_large ?o)))))) |\label{line:4-2}|
    (or                                      |\label{line:5}|
        (forall (?o - ball)
            (or (= ?o ?b) 
                (not (ball_at ?o ?from))))
        (forall (?o - ball)
            (not (ball_at ?o ?to))))         |\label{line:5-2}|
    (forall (?o - ball)                      |\label{line:6}|
            (or ; Implication
                (not (ball_at ?o ?to))
                (or ... ; Ball ?b smaller than ?o
                ))                           |\label{line:6-2}|
 :effect (and
    (occ ?to)                                |\label{line:7}|
    (not (ball_at ?b ?from)) (ball_at ?b ?to)|\label{line:8}|
    (when                                    |\label{line:9}|
        (forall (?o - ball)
            (or (= ?o ?b)
                (not (ball_at ?o ?from))))
        (and (not (char_at ?ppos))
             (char_at ?from)
             (not (occ ?from))))             |\label{line:9-2}|
    (not (snow ?to))                         |\label{line:10}|
    (when                                    |\label{line:11}|
        (and (snow ?to)
             (ball_size_s ?b))
        (and (not (ball_size_s ?b))
             (ball_size_m ?b)))              |\label{line:11-2}|
    (when                                    |\label{line:12}|
        (and (snow ?to)
             (ball_size_m ?b))
        (and (not (ball_size_m ?b))
             (ball_size_l ?b)))              |\label{line:12-2}|
    (increase (total-cost) 1)))              |\label{line:13}|
\end{lstlisting}
  \end{figure}

To perform the action  we first need to ensure that the given three locations are aligned and adjacent (line~\ref{line:1}), ball \texttt{b} is located at the \texttt{from} location (line~\ref{line:2}), and the character is located at the \texttt{ppos} location (line~\ref{line:3}). As the game rules states, we can only move a ball if it does not have any other ball on top, i.e., if it is the smallest ball of the stack (lines~\ref{line:4}--\ref{line:4-2}). Also, if a ball is on top of a stack of balls, it cannot be moved to another stack of balls (lines~\ref{line:5}--\ref{line:5-2}). Finally, to maintain a correct game state, a ball can only be pushed to a stack of balls if it is smaller than all  other balls in the stack (lines~\ref{line:6}--\ref{line:6-2}). 

When the action is performed, since a ball is moved, it modifies the occupancy of some location (lines~\ref{line:7} and~\ref{line:9}--\ref{line:9-2}). Note that occupancy is conditional since the character might be rolling, pushing or popping a ball. We also have to update the ball location (line~\ref{line:8}).
If the character rolls or pushes a ball, its location changes (lines~\ref{line:9}--\ref{line:9-2}).  We remove the snow from the ball location unconditionally (line~\ref{line:10}). If there is snow in the target location, we consider two cases: increasing ball size from small to medium (lines~\ref{line:11}-\ref{line:11-2}) or from medium to large (lines~\ref{line:12}-\ref{line:12-2}). At last, we declare that this action has cost 1 (line~\ref{line:13}).

The \texttt{goal} action (in the single snowman scenarios) has three balls and a location as parameters. Via its precondition it enforces that the three balls are different objects located at the same place. Note that the ball sizes constraint is enforced in the possible state transitions of the \texttt{move\_ball} action.

\medskip

\begin{lstlisting}[escapechar=|, language=PDDL] 
(:action goal
 :parameters 
    (?b0 ?b1 ?b2 - ball ?p0 - loc)
 :precondition (and
    (not (= ?b0 ?b1)) (not (= ?b0 ?b2))
    (not (= ?b1 ?b2))
    (ball_at ?b0 ?p0) (ball_at ?b1 ?p0)
    (ball_at ?b2 ?p0))
 :effect (and  (goal)) )
\end{lstlisting}

\medskip

 As said in the introduction, in Sokoban-like games, sometimes the notion of optimal plans prioritizes box movements minimization.  We propose to also consider this notion of optimality here, hence, correspondingly,  only snow ball movements are considered. This is meaningful in Snowman because, for a human player, it is trivial to identify if there is a path from the current location of the character to the location of the next snow ball movement. 
Therefore, we have the statement \texttt{(:metric minimize (total-cost))} and only the action \texttt{move\_ball} has a cost.

\subsubsection*{Cheating variant}

In order to evaluate how costly is having to find out all the movements of the character, we also evaluate a modified formulation, where the \texttt{move\_character} action is removed and in the \texttt{move\_ball} action it is assumed that the character can teleport (see the empirical evaluation section). Note that this relaxation of the formulation may lead to non-valid plans since sometimes balls will be blocking some paths.


\subsubsection*{Reachability variant}


PDDL offers logical axioms to specify derived predicates, a feature able to cleanly express reachability.
Similarly as done in~\citet{IvankovicH15} with Sokoban, we declare a new predicate \texttt{reachable} with two parameters, the source and target location.
Given a pair of \texttt{?s} (source) and \texttt{?t} (target) locations, the target will be \texttt{reachable} from the source if the target is not occupied and,  it is either the same location as the source  or there exists a middle location \texttt{?m} such that it is next to the source and it is able to reach the target.
\begin{figure}[h!]
\begin{lstlisting}[escapechar=|, language=PDDL]
(:derived (reachable ?s ?t - location)
   (and 
       (or ;; there is a path between ?s and ?t
            (= ?s ?t) ;; either is the same location,
            ;; or it is inductively reachable
            (exists (?d - direction)
                (exists (?m - location)
                    (and
                        (next ?s ?m ?d)
                        (not (occupancy ?m))
                        (reachable ?m ?t)))))
       ;; and target location has no ball in it
       (not (occupancy ?t))))
\end{lstlisting}
\end{figure}

We can now remove the action that moves the character.
We need an additional parameter \texttt{?prevpos} in the \texttt{move\_ball} action, to refer to the position where the character ended in previous state. We then require \texttt{(reachable ?prevpos ?ppos)} in \texttt{move\_ball} action precondition, enforcing that the position where the ball push is going to be executed from is reachable from where the character was. Finally, an additional effect will be required to update the character position, i.e., change the position of the character from \texttt{?prevpos} to \texttt{?ppos}.

\section{Planning as SAT}



Here we recall some basic ideas of Planning as SAT. In this
approach~\cite{kautzS92}, a planning problem is encoded to a Boolean formula,
with the property that any model of this formula will correspond to a valid
plan. 

Since the length of a valid plan is not known a priori, the basic idea is to
encode the existence of a plan of $T$ steps with a formula $f(T)$. Then, the
method for finding the shortest plan consists in iteratively checking the
satisfiability of $f(T)$ for $T = 0,1,2, \dots$ until a satisfiable formula is
found.


Variables need to be replicated for each time step. E.g., $a^{t}$ will denote if
action $a$ is executed at time $t$. Then, the general (standard) encoding goes
as follows. First of all, it is stated that the execution of an action
implies its preconditions, for each $t\in 0..T-1$:
\begin{equation*}
    a^t \rightarrow \Pre^t \text{ for every action } a = \langle \Pre,\Eff \rangle
  \end{equation*}
  
Also, if an action is executed, its effects must take place at the next time step:
\begin{equation*}
    a^t \rightarrow \Eff^{t+1} \text{ for every action } a = \langle \Pre,\Eff \rangle
  \end{equation*}

Preconditions and effects are typically given as sets of literals. Here, by $\Pre^t$ and $\Eff^{t+1}$ we denote the corresponding formulas (usually, conjunctions of literals) on the time-indexed state variables.


Moreover, a change in the value of every state variable $v$ can occur only if an action that can
change this value has been executed:
\begin{equation*}
    (v^t \neq v^{t+1}) \rightarrow \bigvee \{ a^t\mid a= \langle \Pre,\Eff \rangle, v \in \Eff\}
  \end{equation*}

  These are called frame axioms. Finally, it is stated that exactly one action is executed at each time step $t$, and that the goal holds at time $T$.

\section{Snowman SAT encoding}


In this section we propose an encoding for solving the problem at hand, following the
planning as SAT approach.

For clarity and space limitations we do not provide the whole encoding,
but the viewpoint (state variables) and an excerpt of the formulas including the
goal and relevant transition constraints and frame axioms. Also, for simplicity
reasons, we only describe the case of a single snowman.

 %
We represent states with Boolean variables stating, for each location, whether (i) there is snow or not, (ii) there is a ball of a particular size or not, and (iii) there is the character or not. The actions considered are only four, corresponding to a ball movement per possible direction. This will eventually result in rolling, pushing or popping a ball, depending on the current state. In other words, we only need to know the direction of the action.

We consider $L$ to be the set of valid (non-wall)
locations, 
$S$ the set of locations with snow, and $T$ the number of time steps considered. Below we detail the main variables and constraints.

\subsubsection{Variables}
For all $l$ in $L$, $t$ in $0..T$:
  \begin{itemize}
  \item $s_l^t$: there is snow at location $l$ at time $t$
  \item $bs_l^t, bm_l^t, bl_l^t$: there is a small, medium or large ball at location $l$ at time $t$
   \item $c_l^t$: character is at location $l$ at time $t$
  \end{itemize}
For all $t$ in $0..T-1$:
  \begin{itemize}
  \item $n^t, s^t, e^t, w^t$: direction of action at time $t$ is north, south, east or west
  \end{itemize}

  \subsubsection{Constraints}

 \noindent The goal consists in requiring all balls to be at the same location at the end of the plan: 
\begin{gather*}
      \forall l\in L\quad (bs_{l}^T \leftrightarrow  bm_{l}^T \wedge  bm_{l}^T \leftrightarrow bl_{l}^T)
\end{gather*}

\noindent For each time step $t$ in $0..T-1$ we have the following constraints.
\begin{itemize}
  \item Exactly one action is executed per time step: we impose an exactly-one constraint over $n^t, s^t, e^t, w^t$ variables.
  \item Action preconditions and effects (excerpt of the action to move the character north):

Let $L_n$ be the set of valid locations with a wall at north and let $L_{nn}$ be the set of valid locations with a wall two locations ahead at north.

When the character is at any location in $L_n$, it cannot go north:
    \begin{gather*}
      \forall l\in L_n\quad c_l^t\to \neg n^t
    \end{gather*}
Otherwise, if the character moves north and it has a wall two locations ahead, there cannot be any ball in front of him, and at the next time step the location of the character has changed accordingly (here $l_n$ denotes the location at north of $l$):
    \begin{gather*}
      \forall l\in L_{nn}\setminus L_n\quad c_l^t\land n^t\to (move:\\
      \neg c_l^{t+1}\land c_{l_n}^{t+1}\land \neg bs_{l_n}^t\land \neg bm_{l_n}^t\land \neg bl_{l_n}^t)
    \end{gather*}
Finally, if the character moves north without having a wall two locations ahead, apart from moving it can also \emph{roll} a ball north, \emph{push} a ball into a stack of balls or \emph{pop} a ball from a stack of balls. For the sake of brevity we only describe the \emph{push} north action (here $l_{nn}$ denotes the location two steps ahead at north of $l$):
    \begin{gather*}
      \forall l\in L\setminus\{L_n\cup L_{nn}\}\quad c_l^t\land n^t\to (move:\dots\\
      {}\lor push: (\neg c_l^{t+1}\land c_{l_n}^{t+1}\land((\neg bs_{l_n}^{t+1}\land bs_{l_{nn}}^{t+1}\land {}\\
      bs_{l_n}^t\land \neg bm_{l_n}^t\land \neg bl_{l_n}^t\land{}   
      \neg bs_{l_{nn}}^t\land (bm_{l_{nn}}^t\lor bl_{l_{nn}}^t))\\
      {}\lor (\neg bm_{l_n}^{t+1}\land bm_{l_{nn}}^{t+1}\land{}\\
      \neg bs_{l_n}^t\land bm_{l_n}^t\land \neg bl_{l_n}^t\land
      \neg bs_{l_{nn}}^t\land \neg bm_{l_{nn}}^t\land bl_{l_{nn}}^t)))\\
      {}\lor roll: \dots\lor pop: \dots)
    \end{gather*}
  \item Frame axioms impose that the state cannot change without a reason: snow cannot be created, or if it disappears from a location it must be because a ball occupies that location, etc.: 
    \begin{gather*}
      \forall l\in L\quad (\neg s_l^t\to \neg s_l^{t+1})\land{}\\
      (s_l^t\land \neg s_l^{t+1}\to bm_l^{t+1}\lor bl_l^{t+1})\land\dots\\
    \end{gather*}
    
  \end{itemize}


\section{Optimal plans with respect to ball movements}


 As previously done with PDDL, now we also modify the basic encoding by considering only ball movement actions, asking in their preconditions that the location from where the action is performed is reachable from the last location of the character. Additionally, a cheating variant  encoding is included in the experiments for performance comparison. 
  
\subsection{Standard Graph Reachability}


We need to characterize reachability from the location of the character to the
appropriate location next to the ball to be moved. 
\citet{GebserJR14} describe an SMT encoding for \emph{source-target reachability} in graphs, where sources and targets are fixed beforehand. In this section we provide a similar encoding to SAT which is able to deal with unknown a priori sources and targets. We also incorporate the fact that accessible locations may change due to ball movements.
  
The basic idea of the encoding is to prove the existence of a \emph{reachability path} from the source to the target location. We stress that the way of characterising reachability is intrinsically different in SAT or SMT than in PDDL. In particular, we cannot mimic the idea followed in PDDL, where a location $t$ is reachable from $s$ either if $s$ equals $t$ or  there is some (unoccupied) neighbour $s'$ of $s$ such that $t$ is reachable from $s'$. The reason is that, whereas planning adheres to the closed-world assumption (i.e., all unknown values are assumed to be false), this is not the case for SAT and SMT. In other words, the translation of the inductive reachability axioms of PDDL into SAT would be satisfied by any model where all corresponding reachability variables are set to true, hence not encoding reachability at all.

To deal with the open-world assumption of SAT (and SMT) when encoding reachability in graphs, we essentially need to break cyclic relations, i.e., paths from source to target must be encoded as a transitive and antisymmetric relation. As described in \citet{GebserJR14},  acyclicity can be easily modelled with SMT, since an ordering on locations can be imposed by assigning a numeric value to each of them. The details of the encoding are given below and the way we impose  the ordering constraints in SAT is given in next subsection.




We consider a directed graph representation of the maze where the
nodes correspond to the valid locations of the maze, and where there is
a directed edge from $l$ to $l'$ if and only if $l$ and $l'$ are
adjacent locations in the maze. In other words, we define
a directed graph $G=(V,E)$ where $V=L$ and $E=\{(l,l'), (l',l)\mid
l,l' \text{ are adjacent locations in }L\}$. The
encoding goes as follows.

\subsubsection {Variables} We introduce for all $l$ in $L$, $t$ in $0..T-1$, the following
Boolean variable:

\begin{itemize}
\item $r_l^{t}$: node $l$ is reachable by the character at time $t$
\end{itemize}

For a node $l$ to be reachable, either it must be the source or it
must be reached through an edge ending in $l$ and coming from a
reachable node. Therefore, we need variables to state if an edge is
building the reachability path or not. For all edges $(l,l')$ in $E$ and time $t$ we
introduce the following Boolean variable:

\begin{itemize}
\item $e_{l\,l'}^{t}$:  edge $(l,l')$ is in the reachability path at time $t$
\end{itemize}

Moreover,  cycles in those paths must be avoided. Therefore the SMT encoding from~\citet{GebserJR14} would also introduce, for each location $l$ in $L$ and time $t$ the following integer variable:

\begin{itemize}
\item $a_l^{t}\in 1..|L|$: value associated to node $l$ at time $t$; this is used to break cycles: a topological ordering will be enforced among variables $a$ of the nodes in the reachability path to avoid cycles.
\end{itemize}

\subsubsection {Constraints}
The reachability constraints  are the following, where $next(l)= \{l'\mid (l,l')\in E\}$.
\begin{align*}
\forall_{l\in L}\quad & r_l^{t} \rightarrow \big(c_{l}^{t} \bigvee_{l': l \in next(l')}~e_{l'\,l}^{t}\big) \\
\forall_{(l,l')\in E }\quad & e_{l\,l'}^{t} \rightarrow (r_l^{t} \wedge a_l^{t} < a_{l'}^{t})
\end{align*}
Notice that this last ordering constraint is an SMT constraint because it is using the difference logic atom $a^t_l < a^t_{l'}$. We show in next subsection how to replace it in SAT.

\noindent If there is a ball in a location then it is not reachable:
\begin{gather*}
\forall_{l\in L} \quad (bs_l^{t}\lor bm_l^{t}\lor bl_l^{t}) \rightarrow \lnot r_l^{t}
\end{gather*}


\begin{example}
  Consider the graph below. To ease notation, we refer to locations as numbers.

\begin{center}
	\begin{tikzpicture}[auto,node distance= 0.6cm and 0.5cm,>=stealth,baseline={(0,-0.75)}]
	\tikzstyle{nonterminal}=[inner sep= 0pt, font=\scriptsize,circle,thick,draw,minimum size=5mm]
	\tikzstyle{terminal}=[inner sep= 0pt,font=\scriptsize,thick,draw,minimum size=4mm, color = red]
	\tikzstyle{labelfont}=[font=\scriptsize,line width=1.2pt]
          \node[nonterminal] (x1) [] {$1$};
          \node[nonterminal] (x3) [right = of x1] {$3$};
          \node[nonterminal] (x2) [below = of x1] {{$2$}};
          \node[nonterminal] (x4) [right = of x3] {{$4$}};
          \node[nonterminal] (x5) [below = of x3] {$5$};
 	 \node[nonterminal] (x6) [below = of x4] {{$6$}};
 	\path[labelfont] (x1) edge [->,bend right = 30]  (x2);
        \path[labelfont] (x1) edge [->,bend left = 30]  (x3);
        \path[labelfont] (x3) edge [->,bend left = 30]  (x2);
	\path[labelfont] (x3) edge [->,bend left = 30]  (x4);
	\path[labelfont] (x4) edge [ ->,bend left = 30]  (x6);
	\path[labelfont] (x6) edge  [->,bend left = 30] (x4);
	\path[labelfont] (x5) edge  [->,bend right = 30] (x3);
\end{tikzpicture}
\end{center}

  At each time-step, some location $l$ will be asked to be reachable in some action's precondition. In other words, $r_l$ will be asked to be true for some $l$ in $1..6$.

 The encoding for reachability would be the following (we use $b_l$ as a shorthand for $bs_l\lor bm_l\lor bl_l$ and omit time superindexes).

 $$
 \begin{array}{lll}
r_1 \to c_{1}                           & e_{13}\to r_1\wedge a_1<a_3      &   b_1\to \neg{r_1}  \\
r_2 \to c_{2} \vee  e_{12}\vee e_{32} & e_{12}\to r_1 \wedge a_1<a_2    &  b_2\to\neg{r_2} \\
r_3 \to c_{3} \vee  e_{13}\vee e_{53} & e_{32} \to r_3\wedge a_3<a_2    &  b_3\to\neg{r_3}\\
r_4 \to c_{4} \vee  e_{34}\vee {e_{64}} & e_{34} \to r_3\wedge a_3<a_4   & b_4\to\neg{r_4} \\
r_5 \to c_{5}                           & e_{46}\to r_4\wedge a_4<a_6        & b_5\to\neg{r_5}\\
r_6 \to c_{6} \vee e_{46}             &e_{53}\to r_5\wedge a_5<a_3        & b_6\to\neg{r_6}\\
                                      &e_{64}\to r_6\wedge a_6<a_4
 \end{array}
 $$

\end{example}


\subsection{Translation of Ordering Constraints to SAT}

The previous ordering constraints on the values of the numeric variables $a$ associated to locations can be easily translated to SAT. We propose not to encode the numbers to binary form, but to encode the acyclicity relation directly to SAT. 

A \emph{strict partial order} is a relation $<$ that is irreflexive and transitive (which implies antisymmetry as well). This is all we need to ensure acyclicity. We can encode such a relation by adding the constraints
\begin{align*}
  \neg p_{ii}^{t} & \qquad\text{(irreflexivity)}\\
  p_{ij}^{t}\land p_{jk}^{t}\to p_{ik}^{t} & \qquad\text{(transitivity)}
\end{align*}
where $p_{ij}^{t}$ are Boolean variables, for locations $i,j$ and time~$t$. Then, to get rid of SMT constraints and obtain a full SAT encoding, we can simply replace the constraints
\[
\forall_{(l,l')\in E }\quad e_{l\,l'}^{t} \rightarrow (r_l^{t} \wedge a_l^{t} < a_{l'}^{t})
\]
by
\[
  \forall_{(l,l')\in E }\quad e_{l\,l'}^{t} \rightarrow (r_l^{t} \wedge p_{ll'}^{t})
\]
and no numeric variables $a_l^t$ are needed at all.

Notice that transitivity constraints $p_{ij}^{t}\land p_{jk}^{t}\to p_{ik}^{t}$ are only needed for neighbours $j$ of $i$, since transitivity follows by induction. This allows to reduce the number of such constraints from cubic to quadratic.

\subsection{Grid Graph Reachability}

Grids are a particular case of graphs, and a simpler encoding for
reachability in them is possible. The idea is to build a single path
from the source to the target as follows:

define a Boolean variable for each location denoting if it is included in the
path, and impose that the source and destination are in the path and exactly one
of their neighbours is in the path, and that any valid (non-obstacle) location
different from the source and the destination has either zero or two neighbours
in the path.

Not surprisingly, this simple approach is the best performing in our experiments.

\subsection{Invariants}
To help pruning the search space, some simple invariants can be considered.
Since balls cannot decrease their size, the following constraints have to be
fulfilled at any time: the number of large balls cannot exceed the number of
snowmen, and there must be at least as many small balls as snowmen. Imposing
these constraints has shown to be useful only in the harder instances.

Other invariants have been also considered, like forbidding to move balls to locations where they would get stuck without the possibility of building a snowmen, but this does not showed significant improvement.

\section{Empirical Evaluation}\label{sec:evaluation}

For our experiments we used the planners Fast Downward~\cite{fastdownward} v22.12 and SymK \cite{symk} v3.0, and the SAT solver {\sc Kissat}~\cite{kissat} v3.0.0. These two planners were chosen due to their well-known performance and support for the required features present in the PDDL models.
In particular, for Fast Downward we used three configurations: (i) the integrated Stone Soup portfolio~\cite{stonesoup} (ii) LAMA~\cite{lama} and (iii) the blind heuristic.
The Fast Downward Stone Soup portfolio won the satisficing and cost-bounded tracks of the 2018 International Planning Competition (IPC).
When using Fast Downward in preliminary experiments, we observed that all configurations of Stone Soup always resorted to the blind heuristic in our instances, and therefore we only report results for LAMA and the blind heuristic.
As SymK natively supports a variety of PDDL features that are rarely supported by other planners, such as conditional effects and derived predicates with axioms, we used its default bidirectional search.
Finally, the {\sc Kissat} SAT solver and its variations were the winners of various tracks of the 2021 and 2022 SAT competitions~\cite{satcompetition}.
%

To test the efficiency of the proposed approaches we considered all 30 instances of the main part of the game (the real world).\footnote{The game has a second part (the dream world), governed by a different set of rules.} We also included a set of 9 \emph{hidden} instances, that are present in the game files but not used in the actual game. Finally, we crafted 12 instances for some edge cases. More concretely, we considered much bigger scenarios,\footnote{The game scenarios contain at most 40 locations whilst the crafted ones have up to 99 locations.} instances with snow everywhere and some really easy to solve instances with two snowmen. In summary, we considered a total of 51 instances ranging from 1 to 3 snowmen and from 14 to 99 valid locations. 
We ran the experiments on a cluster of compute nodes equipped with Intel Xeon E-2234 CPU @ 3.60GHz processors, where each execution was given a time-limit of 1 hour and 16~GB of memory.

We considered three variations of the PDDL model: basic, cheating and reachability. \emph{Basic} denotes the model with actions able to move the character as well as balls. \emph{Cheating} is a model where we simply remove the action where the character moves, and the character can do ``unsound'' teleports. That is, it does not need to consider if there exists a path between a source and a target position before acting from the target position. Finally, \emph{reachability} denotes a model where derived predicates are used to require  reachability between subsequent action positions in the game. With these three models we aim to evaluate the effect of different encodings of reachability in PDDL.

\begin{table}[h!]
\begin{center}
\begin{tabular}{llr}
\hline
& \textbf{Solved}  & \textbf{PAR-2}  \\
\hline
SAT                & 27               &  196942  \\
SAT-cheating       & 43(19)           &  62414   \\
SAT-R-order        & 43               &  62818   \\
SAT-R-count        & 43               &  60858   \\
SAT-R-count+       & 43               &  60947   \\ \hline
basic-blind        & 27               &  177082  \\
basic-LAMA         & 14               &	275961   \\ 
basic-SymK         & 29               &	166980   \\ \hline
cheating-blind     & 25(16)           &	188086   \\
cheating-LAMA      & 20(11)           &	234919   \\ 
cheating-SymK      & 29(14)           &	162480   \\ \hline
reachability-blind & 26               & 183732   \\
reachability-LAMA  & 16               & 263718   \\ 
reachability-SymK  & 23               & 205981   \\ \hline
\end{tabular}
\caption{
Number of solved instances (within the 1 hour timeout), and PAR-2 scores (the score of a solver is defined as the sum of all runtimes for solved instances + $2\times$timeout for unsolved instances) for each approach. For cheating models the number of valid solutions is given in parenthesis. The total number of instances considered is 51.}
 \label{table:summary}
\end{center}
\end{table}



\begin{figure}[!ht]
\centering
\includegraphics[width=0.47\textwidth]{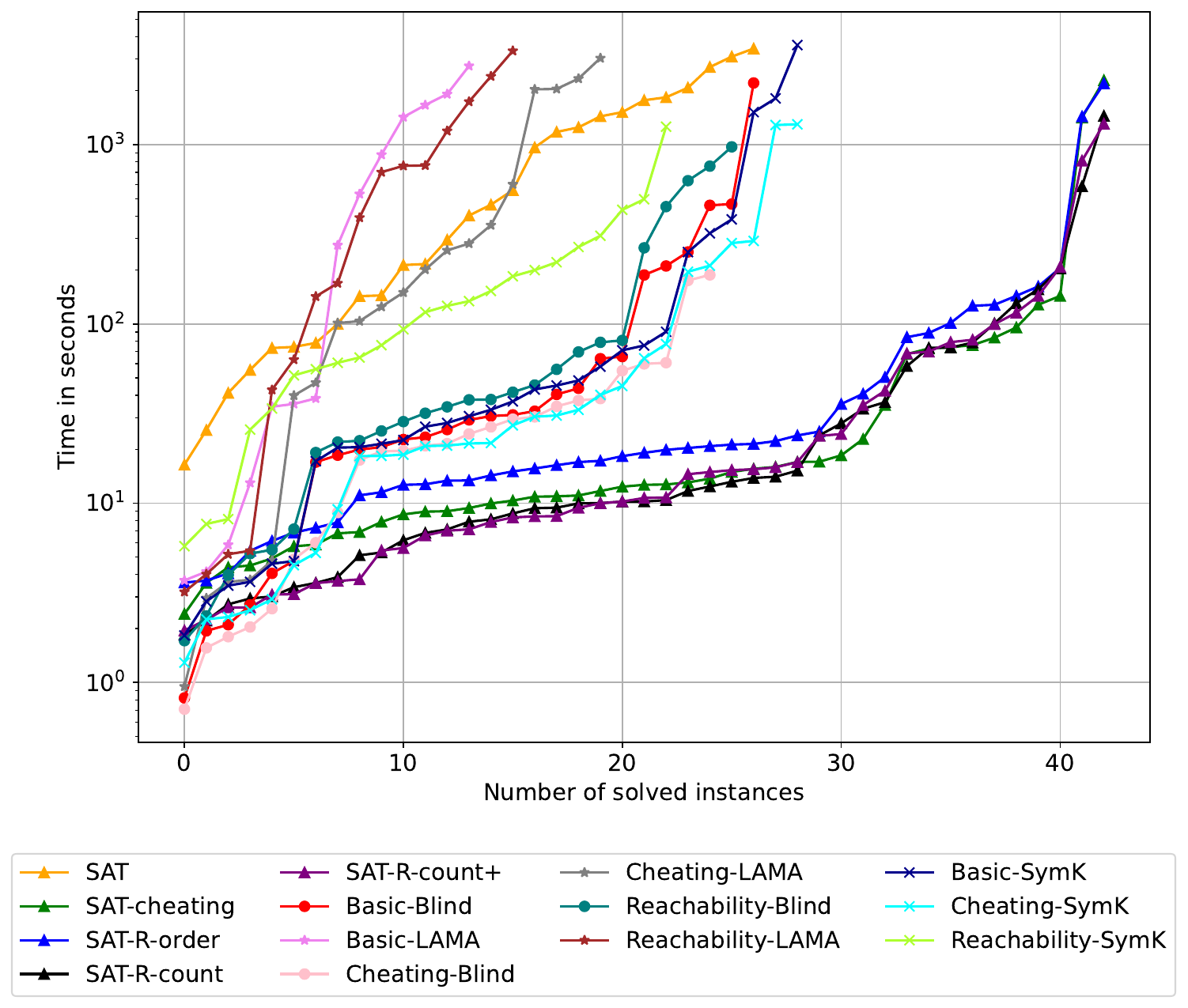}
\caption{Cummulative number of solved instances for all the considered approaches. Time in seconds in logarithmic scale.}\label{fig:cactus}
\end{figure}

Table~\ref{table:summary} summarises all results.\footnote{Detailed results are given in the supplementary material at \url{https://github.com/udg-lai/KEPS2023}.} 
The different SAT configurations are as follows: SAT refers to the basic SAT encoding without reachability constraints, SAT-cheating to a variation of the SAT encoding with ``unsound'' teleports, and the -R versions implement the reachability encodings with the ordering and counting variations, where + denotes the addition of implied constraints~(invariants).
The rest of the rows show the combined performance of each planner and PDDL variant considered. For example, basic-blind considers the basic model solved with Fast Downward using the blind heuristic. 

Column Solved shows the number of instances solved by each system. For the cheating variants, it shows in parenthesis how many solutions are valid out of all the ones solved. Column PAR-2 shows the  sum of all runtimes for solved instances +  2$\times$3600 seconds per timed-out instances.
The cheating and SAT-R variants are clearly the strongest approaches, solving 43 instances out of 51, and showing that the reachability encoding is critical to be able to obtain more (sound) solutions within the given time limit. The cheating variant is not faster than the reachability based ones. Notice that it has less variables but it also forbids less partial assignments, and therefore it may spend more time exploring possible moves. 
This effect may be amplified by the planning as satisfiability approach taken, as for finding a length-optimal plan it previously proves unfeasibility of all shorter plans.

When comparing the cheating and basic PDDL models, the cheating variant only results into notable gains with LAMA. When comparing the cheating model with the reachability model, there is a notable difference in both LAMA and SymK. Although not visible in the summarized results, amongst the PDDL approaches with two or three snowmen, only SymK was able to solve two of them within the given time limit. The rest of the planners were not able to preprocess any of them due to the increased complexity of the domain.

Following the PAR-2 scores, three main groups can be observed:  the SAT-R variants as the best performers, SAT, blind search and SymK  as the second group, and finally LAMA being notably slower. 
The implied constraints (invariants) did not seem to help in reducing the solving time, at least for the instances considered.
Figure~\ref{fig:cactus} depicts the cummulative number of solved instances over time.  Note the time axis is in logarithmic scale. This figure allows to identify more clearly how fast are the three aforementioned groups. 

\section{Conclusions and Future Work}

In this work we contributed new challenging planning benchmarks.
%
In the framework of puzzle-like games, we believe that SAT models like the ones we have presented could be used to assist in the (semi)automatic design of new scenarios, for instance, by ensuring that randomly generated or crafted scenarios satisfy the desired constraints in terms of difficulty of the game levels.

We have shown how relatively simple encodings to SAT outperform by far state-of-the-art planners in the game \emph{A Good Snowman is Hard to Build}, especially when considering reachability constraints.

Some works have introduced SAT and SMT solvers with support for detecting acyclicity and reachability in graphs~\cite{GebserJR14,DBLP:conf/aaai/BaylessBHH15}. Therefore, an alternative approach could be to use a SAT or SMT solver with built-in support for reachability, such as {\sc MonoSAT}~\cite{DBLP:conf/aaai/BaylessBHH15}.

As future work, we plan to address the simultaneous (parallel) execution of actions as an additional way of reducing the time span, in order to overcome scalability issues and be able to solve the harder instances in a reasonable amount of time. This will require non-trivial checking of possible interferences between actions.





\section*{Acknowledgements}
Work partially supported by grant PID2021-122274OB-I00 funded by MCIN/AEI/10.13039/501100011033 and by ERDF A way of making Europe.

\bibliography{main}

\end{document}